\def\year{2020}\relax
\documentclass[letterpaper]{article} 
\usepackage{aaai20}  
\usepackage{times}  
\usepackage{helvet} 
\usepackage{courier}  
\usepackage[hyphens]{url}  
\usepackage{amsfonts}
\usepackage{amsmath}
\usepackage{bm}
\usepackage{graphicx} 
\usepackage{graphicx}  
\title{Cross-Lingual Low-Resource Set-to-Description Retrieval \\for Global E-Commerce}
\author{Juntao Li,\textsuperscript{\rm 1,2}$^{\ast}$ Chang Liu,\textsuperscript{\rm 1,2}\thanks{Equal contribution} Jian Wang,\textsuperscript{\rm 3} Lidong Bing, \textsuperscript{\rm 3}\\
\Large \textbf {Hongsong Li,\textsuperscript{\rm 3} Xiaozhong Liu,\textsuperscript{\rm 4} Dongyan Zhao,\textsuperscript{\rm 1,2} Rui Yan\textsuperscript{\rm 1,2}\thanks{Corresponding author: Rui Yan (ruiyan@pku.edu.cn)}}\\
$^1$Center for Data Science, Academy for \\
Advanced Interdisciplinary Studies,
Peking University, Beijing, China\\
 $^2$Wangxuan Institute of Computer Technology, Peking University, Beijing, China\\
 $^3$ DAMO Academy, Alibaba Group\\
 $^4$ Indiana University, Bloomington, USA\\
 \tt \{lijuntao,liuchang97,zhaody,ruiyan\}@pku.edu.cn, \tt binglidong@gmail.com\\
\tt \{eric.wj,hongsong.lhs\}@alibaba-inc.com, \tt liu237@indiana.edu
}
 
\begin{document}

\maketitle

\begin{abstract}
With the prosperous of cross-border e-commerce, there is an urgent demand for designing intelligent approaches for assisting e-commerce sellers to offer local products for consumers from all over the world.
In this paper, we explore a new task of cross-lingual information retrieval, i.e., cross-lingual set-to-description retrieval in cross-border e-commerce, which involves matching product attribute sets in the source language with persuasive product descriptions in the target language.
We manually collect a new and high-quality paired dataset, where each pair contains an unordered product attribute set in the source language and an informative product description in the target language.
As the dataset construction process is both time-consuming and costly, the new dataset only comprises of 13.5k pairs, which is a low-resource setting and can be viewed as a challenging testbed for model development and evaluation in cross-border e-commerce.
To tackle this cross-lingual set-to-description retrieval task, we propose a novel cross-lingual matching network (CLMN) with the enhancement of context-dependent cross-lingual mapping upon the pre-trained monolingual BERT representations.
Experimental results indicate that our proposed CLMN yields impressive results on the challenging task and the context-dependent cross-lingual mapping on BERT yields noticeable improvement over the pre-trained multi-lingual BERT model.
\end{abstract}

 \section{Introduction}

Cross-border online shopping has become popular in the past few years, and e-commerce platforms start to pay more effort in improving their user experience.
To increase the conversion rate, one critical aspect is how to display products to overseas customers with persuasive and informative descriptions.
Fortunately, there already exist various local and cross-national e-commerce platforms, e.g., Amazon, eBay, Taobao, Lazada, with multi-lingual persuasive descriptions of products from all over the world, making it possible for retrieving a feasible product description in the foreign language to a local product.
Compared with generation-based methods, retrieved product descriptions are more understandable and accessible to overseas users, and training a retrieval model requires less paired data.
Accordingly, we formulate this problem as a cross-lingual information retrieval (CLIR) task for global e-commerce, i.e., ranking foreign product descriptions against a local product with an attribute set \cite{litschko2018unsupervised,zhang2019improving}.

\begin{figure}[t]
\centering
\includegraphics[width=1\columnwidth]{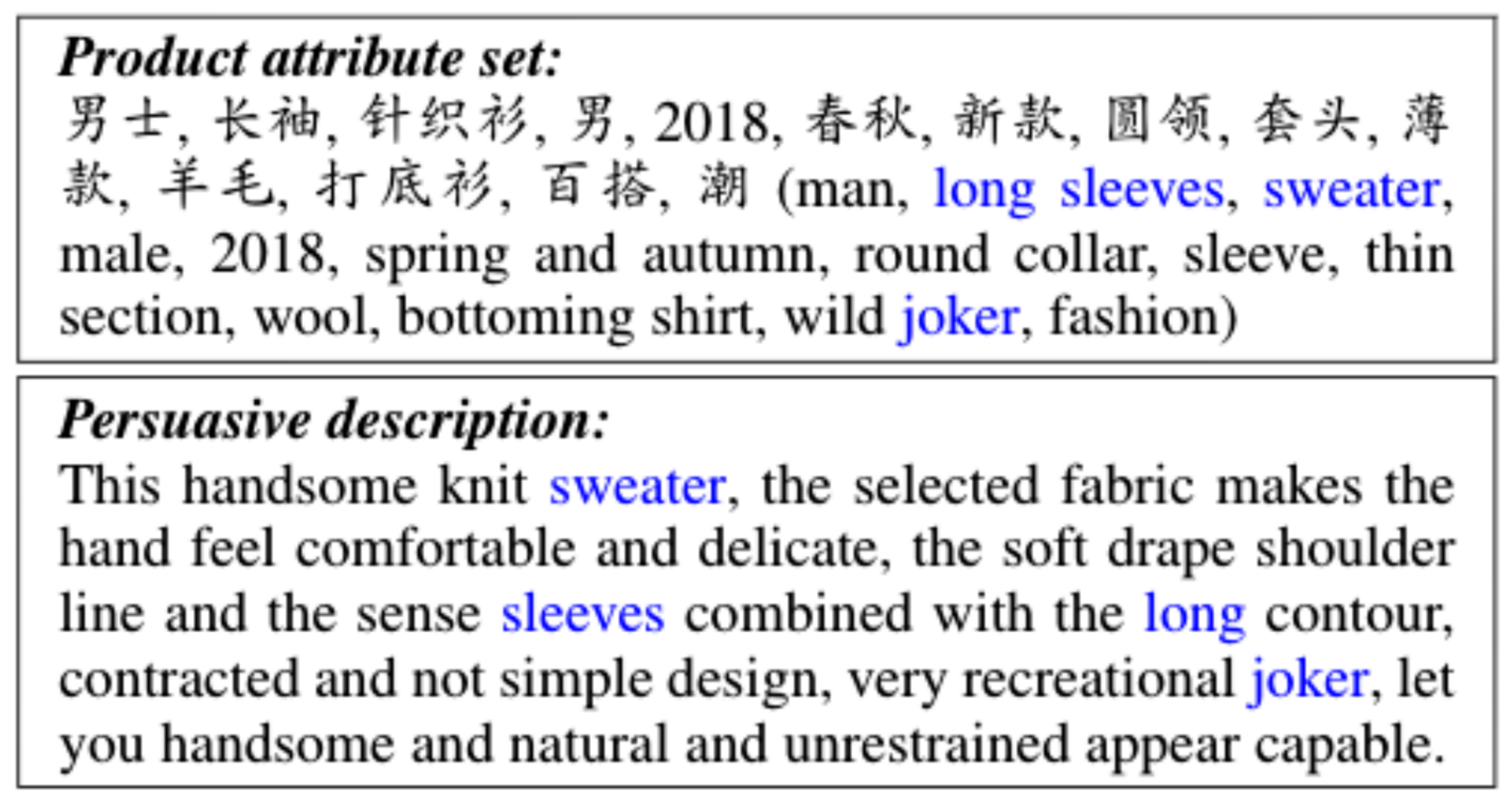}
\caption{An example training pair sampled from the collected dataset, where a Chinese product attribute set is given for retrieving a proper persuasive description in English.}
\label{fig:running_sample}
\end{figure}

Conventional CLIR systems mainly comprise of two components: machine translation and monolingual information retrieval.
According to the translation direction, these systems are further categorized into translating the local language to the foreign one and translating the foreign language to the local one.
By doing so, the CLIR task is converted to a monolingual setting through the machine translation system \cite{nie2010cross,ruckle2019improved}.
Although these systems are conceptual simple and natural, they are restricted by the performance of machine translation system.
Thus, directly modeling CLIR with recent deep neural information retrieval models is promising \cite{hui2018co,mcdonald2018deep}. 
However, while performing well, existing deep IR models require a large amount of labeled training data, which is expensive and costly in our cross-lingual retrieval scenario.

In this paper, we explore to leverage a deep neural model to directly solve the low-resource cross-lingual information retrieval task without machine translation process. 
Herein, the critical component of deep IR model is to capture the matching score between product attribute sets in the source language and product descriptions in the target language.
As there is not existing open dataset for CLIR in e-commerce,
we hire domain experts to collect a heuristic dataset with 13.5K pairs to facilitate the development of cross-lingual retrieval in e-commerce and address the shortage of dataset.
Note that since present retrieval/searching systems of cross-border e-commerce platform are not yet applicable to obtain product description candidates in the target language for re-ranking, we mainly focus on designing matching model to solve this low-resource cross-lingual problem directly.
Figure \ref{fig:running_sample} illustrates an example in the collected dataset, where an attribute set (i.e., keywords) of a product in the source language has a matched persuasive description in the target language.
Specifically, the product attribute sets are collected from a local e-commerce platform, where there exist various overlapping and relations inside the sets to improve the clicking rate.
The corresponded heuristic product descriptions in the target language are created by domain experts, and only leverage part of the attribute set considering the overlapping and relations inside the set.

To address the obstacle of lacking paired data and the unordered product attribute sets, we propose a novel model with the enhancement of unsupervised monolingual pre-training, semi-supervised cross-lingual alignment, and a supervised fine-tuning matching model. 
Concretely, we propose to leverage the deep attention module in neural translation \cite{vaswani2017attention} and response selection \cite{zhou2018multi} for modeling the unordered product attributes and copious cross-lingual correlations between the matched pairs in two different languages.
Besides, motivated by the appealing performance of unsupervised monolingual pre-trained models (e.g., ELMo \cite{peters2018deep}, BERT \cite{devlin2018bert}), we further incorporate a semi-supervised context-dependent cross-lingual mapping mechanism upon the monolingual BERT representations.
By doing so, the data scarcity issue is significantly mitigated.

In a nutshell, our contributions are as follows. 
(1) We gathered a new and challenging heuristic dataset for cross-lingual retrieval in e-commerce. 
The dataset will be given by asking.
(2) We proposed a cross-lingual matching network (CLMN) for addressing this task, which incorporates deep attention mechanism for modeling unordered attribute sets and cross-lingual correlations, and a context-dependent cross-lingual mapping learning upon BERT. 
The code is available on \url{https://github.com/LiuChang97/CLMN}.
(3) We conducted extensive experiments on the challenging dataset, including evaluating the performance of fine-tuned multi-lingual BERT representations, exploring the performance of strong machine translation based CLIR methods, comparing with the state-of-the-art directly CLIR models. 
(4) Experimental results indicate that our proposed CLMN model achieves promising performance on the low-resource cross-lingual retrieval task.
With the enhancement of context-dependent mapping upon BERT, our directly cross-lingual retrieval model achieves significant performance improvement over the fine-tuned Multi-lingual BERT and state-of-the-art CLIR models.

\section{Related Work}
\subsection{Information Retrieval Models}
Information retrieval models for cross-lingual tasks can be roughly categorized into two groups. 
The first one is to learn matching signals across different language space directly.
For instance, bag-of-words addition has been proved effective \cite{vulic2015monolingual} for matching degree calculation, where each text is represented by adding its bilingual word embeddings representation together. 
With the enhancement of the TF-IDF algorithm, bag-of-words addition model further achieves a performance improvement, and the term-by-term query translation model yields the state-of-the-art performance for unsupervised cross-lingual retrieval \cite{litschko2018unsupervised}.
Another group of methods transfer the cross-lingual retrieval task to a monolingual retrieval task by using machine translation systems \cite{schuster2019cross}, e.g., combing cross-language tree kernel align with neural machine translation system for retrieval \cite{da2017cross}, learning language invariant representations for cross-lingual question re-ranking \cite{joty2017cross}.

\subsection{Low-Resource Cross-Lingual Learning}
Existing low-resource cross-lingual learning methods mainly contain three elements, i.e., off-the-shelf monolingual word embedding training algorithm \cite{bojanowski2017enriching}, cross-lingual mapping learning, and the refinement procedure.
In cross-lingual mapping learning, a linear mapping between the source embeddings space and the target embeddings space is learned in an adversarial fashion \cite{conneau2017word}.
To enhance the quality of learned bilingual word embeddings, various refinement strategies are proposed, such as synthetic parallel vocabulary building \cite{artetxe2017learning}, orthogonal constraint \cite{smith2017offline}, cross-domain similarity local
scaling \cite{sogaard2018limitations}, self-boosting \cite{artetxe2018robust}, byte-pair encodings \cite{sennrich2016neural,lample2018phrase}.
Alternatively, a context-dependent cross-lingual representation mapping based on pre-trained ELMo \cite{peters2018deep} is proposed recently to boost the performance of cross-lingual learning.
\cite{schuster2019cross}.
Unlike previous work, we propose to train a cross-lingual mapping upon the context-dependent monolingual BERT with an effective refinement strategy. 

\begin{figure*}[t]
\centering
\includegraphics[width=1\textwidth]{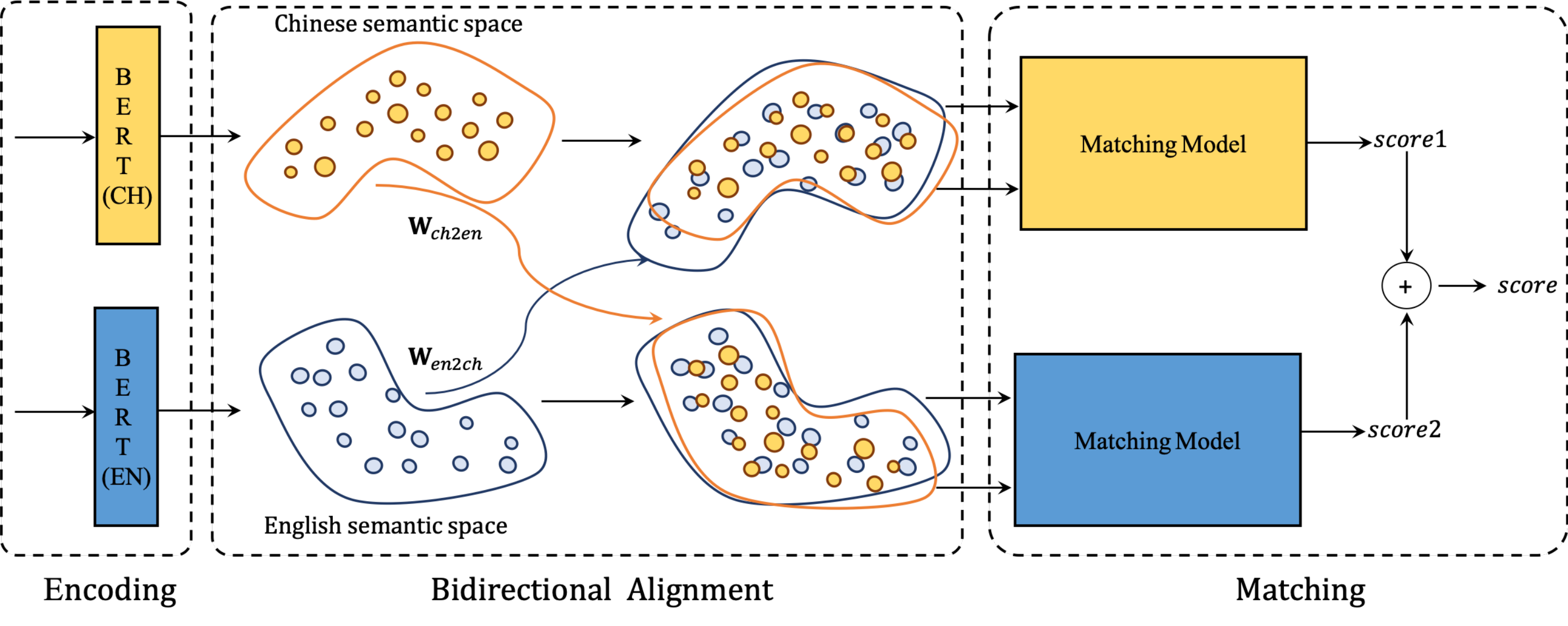}
\caption{\label{fig:framework}The overall architecture of our proposed CLMN model.}
\end{figure*}

\subsection{Existing Datasets}
There already exists several cross-lingual text retrieval datasets.
Most of them are collected from open-domain and comprise of comparable/parallel document pairs, e.g., WaCky translation dataset \cite{baroni2009wacky,joulin2018loss}, benchmark CLEF corpora \cite{vulic2015monolingual}, the Askubuntu benchmark corpus in QA task \cite{dos2015learning,barzilay2016semi}, Arabic–English language pairs \cite{da2017cross}, text stream alignment \cite{ge2018fine}, English-German cross-lingual information retrieval dataset based on English-German Wikipidia \cite{schamoni2014learning}.
Later on, Sasaki et al. \shortcite{sasaki2018cross}  follow previous dataset construction method \cite{schamoni2014learning} and collect 25 cross-lingual datasets with large scales based on Wikipedia.
R{\"u}ckl{\'e} et al. \shortcite{ruckle2019improved} extend open-domain cross-lingual question retrieval to the task-oriented domain, i.e., constructing a dataset upon StackOverflow.
To facilitate the development of cross-lingual information retrieval in cross-border e-Commerce, we, for the first time, create a high-quality heuristic dataset from real commercial applications.

\section{Our Proposed Model}

In our cross-lingual set-to-description retrieval task, the main purpose is to train a cross-lingual matching model that computes the relevance score between a product attribute set in the source language and a product description in the target language.
To formulate this task, we utilize the following necessary notations.
A dataset with paired product attribute sets and descriptions
\begin{math}
\mathcal{D}=\{(a_i, d_i, y_i)\}_{i=1}^N
\end{math} 
is first given, where $a_i$, $d_i$, $y_i$ represent a product attribute set in the source language, a description in the target language, and the corresponding relevance score $y_i \in \{0,1\}$ where $y_i$ = 1 represents the given description candidate is a proper one for the given attribute set, otherwise $y_i=0$.
Our task is defined as learning a matching model from the given dataset that can output a relevance score for a set-description pair.
As shown in Figure \ref{fig:framework}, our proposed cross-lingual matching network (CLMN) consists of encoding, bidirectional cross-lingual alignment, and matching.

\subsection{Encoding}
As the unsupervised monolingual pre-trained models has achieved promising results on various NLU tasks \cite{peters2018deep,devlin2018bert}, we propose to leverage the pre-trained BERT for obtaining the contextualized word representations.
In details, we utilize the pre-trained Chinese BERT and the pre-trained English BERT to convert each product attribute set(Chinese) and each product description(English) into two separate monolingual semantic spaces.
Their monolingual contextualized representations are then bidirectionally aligned into the same semantic spaces by context-dependent bilingual mapping.

\begin{figure*}[t]
\centering
\includegraphics[width=1\textwidth]{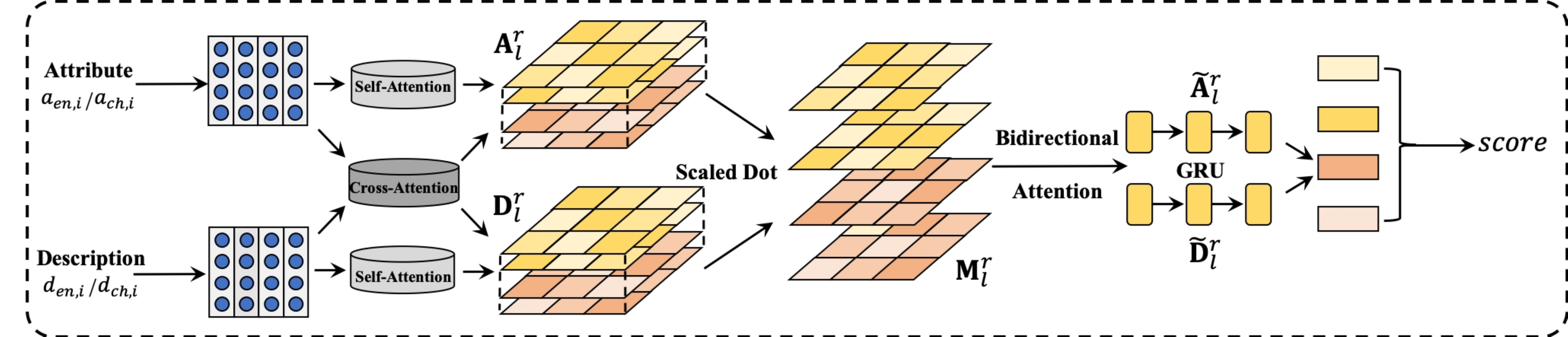}
\caption{\label{fig:matching}The detailed illustration of the matching model.}
\end{figure*}

\subsection{Bidirectional Cross-lingual Alignment}
Inspired by the success of cross-lingual alignment learning \cite{conneau2017word}, we propose to map the contextualized monolingual embedding representations to a shared bilingual space so as to extract semantic information and matching features of a product attribute set and a description.
Different from the case in context-independent word embeddings, the contextualized embedding of a word varies with the surrounding contexts. 
In order to draw on the effective off-the-shelf cross-lingual alignment algorithms, we bridge this gap by considering the averaged contextualized representations as the anchor(i.e., the static embedding) for each word in a monolingual corpus following Schuster et al. \shortcite{schuster2019cross}. Then we adopt MUSE \cite{conneau2017word} to learn the cross-lingual alignment matrix $\bf W$ as the initialization of the mapping matrix of our model.
To better facilitate the downstream matching task, we propose to update $\bf W$ during the training process. 
As demonstrated by Smith et al. \shortcite{smith2017offline}, imposing an orthogonal constraint to the mapping matrix leads to better performance.
In order to ensure that $\bf W$ stays close to an orthogonal matrix, we put an additional constraint during parameters updating~\cite{cisse2017parseval,conneau2017word}, denoted as:
\begin{equation}
    \bf W \leftarrow (\text{1} +\beta)\bf W - \beta(\bf W\cdot \bf W^\mathrm{T})\cdot \bf W
\end{equation}

As illustrated in Figure \ref{fig:framework}, we conduct bidirectional cross-lingual alignment on the contextualized monolingual embeddings.
We first learn the alignment mapping matrix $\bf W$$_{ch2en}$ from the Chinese space to the English Space.
Then, we learn the alignment mapping matrix $\bf W$$_{en2ch}$ from the English space to the Chinese Space.
By doing so, we have two groups of representations of $a_i$ and $d_i$ in the bilingual space, i.e., $a_{en,i}$ and $d_{en,i}$ in the English semantic space and $a_{ch,i}$ and $d_{ch,i}$ in the Chinese semantic space.

\subsection{Matching Model}

Given that the product attribute sets are unordered and there exist abundant correlations between product attributes and descriptions, we propose to use a fully attention-based module, including self-attention and cross-attention, to capture these dependency information.
Specifically, the attention-based module is constructed upon the aligned bilingual representations of each product attribute and description.
The attentive module takes three inputs, namely the query, the key, and the value, which are denoted as ${\bf {Q}}$, ${\bf {K}}$, ${\bf {V}}$ respectively.
First, the attentive module uses each word in the query to attend each word in the key through the scaled dot-product attention mechanism.
Then, the obtained attention score is functioned upon the value $\bf {V}$ to form a new representation of $\bf {Q}$, which is formulated as
\begin{equation}\label{eq:att}
\textit{Att}({\bf {Q,K,V}})= \textit{softmax}(\frac{{\bf {Q\cdot K}^\mathrm{T}}}{\sqrt{d}})\cdot \bf {V}
\end{equation}
Thus, each word in the query $\bf {Q}$ is represented by the joint meaning of its similar words in $\bf {V}$.
In practice, the key $\bf {K}$ and the value $\bf {V}$ are set to identical.
$\bf {Q}=\bf {V}$ corresponds to the self-attention representation of the query. Otherwise, we can obtain the cross-attention representation of the query.
We stack the attention modules $L$ times to better capturing different granularities of representations \cite{zhou2018multi}. 
Accordingly, we obtain the self-attention and the cross-attention representations $\mathbf A_{l}^r$ and $\mathbf D_{l}^r$, where $1 \leq l \leq L$ and $r \in \{self, cross\}$.
To perform interaction between two sets of representations, we first calculate the scaled interaction matrix $\mathbf M_l^r$, denoted as:
\begin{equation}
    \mathbf M_{l}^r = \frac{\mathbf A_{l}^r \cdot {\mathbf{D}_{l}^r}^\mathrm{T}}{\sqrt{d}} 
\end{equation}
We then use softmax function on both row and column directions of $\mathbf M_{k}^r$ to get bi-directional attentive weights and obtain the bi-directional attentive aligned representations:
\begin{equation}
\begin{aligned}
& \mathbf{\hat A}_{l}^r = \textit{softmax}(\mathbf M_{l}^r) \cdot {\mathbf{D}_{l}^r} \\
	& \mathbf{\hat D}_{l}^r = \textit{softmax}({\mathbf{M}_{l}^r}^\mathrm{T}) \cdot {\mathbf{A}_{l}^r} \\
\end{aligned}
\end{equation}
Next, we calculate the interaction between $\mathbf A_{l}^r$ and $\mathbf{\hat A}_{l}^r$ as well as $\mathbf D_{l}^r$ and $\mathbf{\hat D}_{l}^r$ by element-wise multiplication, subtraction, and concatenation, following with a fully connected layer with ReLU activation denoted as $g(\cdot)$:
\begin{equation}
\begin{aligned}
	& \mathbf{\tilde A}_l^r = g([\mathbf A_l^r \odot \mathbf{\hat A}_l^r;\mathbf A_l^r - \mathbf{\hat A}_l^r;\mathbf A_l^r;\mathbf{\hat A}_l^r])\\
	& \mathbf{\tilde D}_l^r = g([\mathbf D_l^r \odot \mathbf{\hat D}_l^r;\mathbf D_l^r- \mathbf{\hat D}_l^r;\mathbf D_l^r;\mathbf{\hat D}_l^r])\\
\end{aligned}
\end{equation}
Subsequently, we consider the rows of $\mathbf{\tilde A}_l^r$ and $\mathbf{\tilde D}_l^r$ as time steps and summarize the interaction information with two GRU layers then concatenate the last hidden states of both as the representation of the overall interaction features of $\mathbf A_{l}^r$ and $\mathbf D_{l}^r$.
Finally, we gather all the interaction information from all the interaction between all the $\mathbf A_l^r$ - $\mathbf D_l^r$ pairs with a fully connected layer followed by a sigmoid function to obtain the matching scores. We denote the predictions of the matching models in the Chinese semantic space and the English semantic space $score1$ and $score2$, respectively.

\subsection{Learning and Prediction}
Recall that we have two separate matching models with the same structure but different inputs from different aligned semantic spaces, we jointly update the two matching models as well as the two alignment matrices with the same batches of training cases so as to make full use of the forward pass of BERT, which is time-consuming.
In learning each of the matching model $f(\cdot)$, the objective is to minimize the cross entropy with dataset $\mathcal{D}$, formulated as:
\begin{equation}
\begin{aligned}
\mathcal{L}=&-\sum_{i=1}^Ny_ilog(f(a_i,d_i,y_i)) \\
&+ \sum_{i=1}^N(1-y_i)log(1-f(a_i,d_i,y_i))
\end{aligned}
\end{equation}
where $N$ refers to the number of training pairs, $y_i$ is the label of a product descriptions, $f(\cdot)$ is the matching model.

In prediction, we utilize the addition of $score1$ and $score2$ as the final relevance score between product attribute sets and descriptions.

\section{Experimental Setup}

\subsection{CLIR-EC Dataset}

We propose to construct a heuristic dataset CLIR-EC (Cross-Lingual Information Retrieval for e-Commerce) for exploring how to retrieve informative product descriptions in the target language given product attribute sets in the source language.
Specifically, we hired 20 annotators with both proficient language skills (with master's degree majored in translation) and domain knowledge (at least one year experience for e-commerce data annotation) to write a proper informative product description in the target language given a product attribute set in the source language.
As there exist multi-level overlaps and correlations inside a product attribute, we asked annotators to choose part of the pivotal attributes of a set for creating a fluent and persuasive product description.
We also employed 3 native speakers to conduct proofreading and editing on the collected product descriptions.
By doing so, each product attribute set in the source language is paired with a product description in the target language, which is used for training the CLMN model. 
\begin{table}[!t]
\centering
\resizebox{\columnwidth}{!}{
\begin{tabular}{|l| c c | c c|}
\hline
 & \bf Taobao  & \bf eBay  & \bf Attri. &\bf Descri. \\
 \hline
\# Samples &  313K &  118K  &  13.5K & 13.5K\\
\# Total Words &  17.6M &  8.92M  &  233K & 639K\\
Average Length  &  56.2  & 75.8  &  17.2 & 47.3\\
Vocabulary Size  & 163K  & 50.8K  &  16.6K & 15.5K\\
\hline
\end{tabular}%
}
\caption{\label{tab:data}  The statistical results of datasets used in experiments. \textbf{Descri.} refers to product descriptions. \textbf{Attri.} is product attribute sets.}
\end{table}
Totally, we collected 13.5 K training pairs, and the collection process takes 35 days.
More details can be seen in Table \ref{tab:data}.
We randomly split the alignment dataset into 10,500/1,000/2,000 for training/validating/testing the CLMN model.
We also launch another human evaluation to verify the quality of the manually created dataset.
Concretely, we employed 5 evaluators with proficient language skills and rich cross-national shopping experience to judge the created product descriptions from the following four perspectives: 1) \textbf{information integrity}, i.e., whether the descriptions include the most pivotal information in product attributes; 2) \textbf{grammatically formed}, i.e., whether the descriptions are fluent and grammatical; 3) \textbf{matching degree}, i.e., does the description correlate well with any given attributes; 4) \textbf{sentiment polarities}, i.e., whether the descriptions present a positive sentiment to users.
Each annotator is asked to rate a product description from these aspects on a scale of 1 (very bad) to 5 (excellent).
We randomly sample 500 pairs for quality verification.
Table \ref{tab:quality} presents the results of human verification, where the evaluation results are very promising.

\subsection{Other Datasets}
Besides, we also automatically collect two large monolingual corpora from e-commerce platforms, i.e., eBay \footnote{https://www.ebay.com/} with product descriptions in the target language (English), and Taobao \footnote{https://www.taobao.com/} with product attribute sets in the source language (Chinese) for learning cross-lingual alignment, including bilingual word embeddings and context-dependent bilingual word representations upon BERT.
Table \ref{tab:data} illustrates the detailed data statistical results.
Overall, there are 313K samples and 17.6 million words in the Taobao dataset while there are 118K samples and 8.92 million words in the eBay dataset.
The average document length of Taobao and eBay corpora are 75.8 and 40.8, respectively.
To preprocess these corpora, we use NLTK for tokenization. 
All words in both datasets are reserved, which results in a vocabulary size of 162,506 for Taobao and 59,233 for eBay.

\begin{table}[!t]
\centering
\small
\begin{tabular}{|l|c c |}
\hline
\bf Metrics                & \bf Scores   & \bf Kappa Values\\
\hline
Grammatically Formed                & 4.77       & 0.71      \\
Information Integrity     & 4.58       & 0.63      \\
Sentimental Polarities     & 4.79       & 0.80      \\
Matching Degree & 4.53       & 0.75      \\
\hline
\end{tabular}
\caption{\label{tab:quality}Evaluation results of data quality verification, where kappa values refer to the inter-annotator agreement.}
\end{table}

\subsection{Comparison Methods}
To thoroughly evaluate the performance of our proposed CLMN model, we leverage three groups of baselines, i.e., advanced unsupervised cross-lingual IR models, the combination of machine translation system and monolingual IR models, directly cross-lingual retrieval models.

\subsubsection{Unsupervised CLIR Models} We utilize the state-of-the-art unsupervised cross-lingual retrieval model and its extensions \cite{litschko2018unsupervised}, including term-by-term query translation model (\textbf{TbTQT}), bilingual word embedding aggregation model (\textbf{BWE-AGG}), inverse document frequency weighted BWE-AGG (\textbf{BWE-IDF}), \textbf{Bigram matching}.

\subsubsection{Translation Based CLIR Models} 
We follow the recent proposed CLIR model \cite{ruckle2019improved} to transfer the cross-lingual information retrieval task to a monolingual information retrieval one by a machine translation model. In our settings, we use the SOTA commercial translation system (Google API) to translate the unordered product attribute sets in the source language to the target language, and then we utilize the obtained paired data to train the following monolingual IR models for retrieval.

\textbf{SMN}, a strong model for response selection~\cite{wu2017sequential}. It first learns a matching vector between the translated attribute sets and the product descriptions with the CNN network. Then, the learned matching vectors are aggregated by an RNN to calculate the final matching score.

\textbf{DAM}, a strong text matching network with only attention modules, which is adapted from the response selection task~\cite{zhou2018multi}. This model is used for studying whether attention modules can model unordered attribute sets and their matching degrees with descriptions.

\textbf{CSRAN}~\cite{tay2018co}, a state-of-the-art monolingual IR model. This model first adopts stacked recurrent encoders with refinement to learn representations. Then, it calculates interactions based on the attentively aligned representations and aggregates the interaction features via a multi-layered LSTM to obtain matching scores.

\subsubsection{Directly CLIR Models}
We also leverage the state-of-the-art directly cross-lingual IR models for comparison.

\textbf{ML-BERT}~\cite{devlin2018bert}, the multilingual version of BERT \footnote{https://github.com/google-research/bert}, which is trained on 104 languages and has achieved the SOTA performance in cross-lingual text pair classification \cite{conneau2018xnli}. We further fine-tune the pre-trained multilingual BERT on our collected datasets.

\textbf{POSIT-DRMM} \cite{zhang2019improving}, the recent proposed cross-lingual document retrieval model, which is designed for addressing the low-resource issue in CLIR. 
This model incorporates bilingual representations to capture and aggregate matching signals between an input query in the source language and a document in the target language.

\subsection{Implementation Details}

Following the conventional settings in related research \cite{zhou2018multi} for training the proposed CLMN model, we use the CLIR-EC dataset to prepare the supervised positive sample pairs and negative sample pairs.
Each positive sample pair consists of a product attribute set and the matched product description while a negative sample pair comprises of a product attribute set and a mismatched product description sampled from all of the descriptions in CLIR-EC dataset.
The proportion of the positive sample and the negative sample in training and validation sets is 1:1, while the corresponding proportion of testing set is 1:9.
We limit the length of product attributes and descriptions to 50 words and 100 words, respectively.
The pre-trained static monolingual and bilingual word embeddings on Taobao and eBay corpora used by all the non-BERT models have the dimension size 300.  
For our BERT-based CLMN model, we use the 768-dimension outputs of the second last BERT layer and keep the dimension of the encoder outputs as well as interaction outputs the same as the dimension of BERT.
We use $L$ = 2 stacked attention modules for attributes and descriptions. 
The parameters of encoders for attributes and descriptions are shared.
We set the mini-batch size to 50 and adopt Adam optimizer \cite{kingma2014adam} with the initial learning rate of 3e-4 to for training.
The best performance of each model on the validation set is chosen.
\subsection{Evaluation Metrics}
To evaluate the model performance, we follow the conventional settings in related work \cite{wu2017sequential,zhou2018multi,zhang2019improving}.
Specifically, we first calculate the matching scores between a product attribute set and product description candidates, and then rank the matching scores of all candidates to calculate the following automatic metrics, including mean reciprocal rank (MRR) \cite{voorhees1999trec}, and recall at position k in n candidates (Rn@k).

\begin{table}[!t]
\centering
\large
\resizebox{\columnwidth}{!}{
\begin{tabular}{|l c c c c c|}
\hline
\textbf{Model}   & \bm{$R_2@1$}  & \bm{$R_{10}@1$}  &\bm{$R_{10}@2$} &\bm{$R_{10}@5$} &\bm{$MRR$}\\
\hline
\multicolumn{6}{l}{Unsupervised models}\\
\hline 
TbTQT &  86.0 &  40.1 & 64.7 &  94.5 & 61.7\\
BWE-AGG  &64.1 & 20.4 & 33.6 &65.5  &40.3 \\
BWE-IDF &62.0 &  19.0 & 32.6 & 66.4 &39.3\\
Bigram &64.0 &19.4 &  33.4 & 65.3 & 39.6\\
\hline
\multicolumn{6}{l}{Translation+Monolingual information retrieval models}\\
\hline 
SMN &94.7 &73.6 &  88.8 & 98.4 & 84.2\\
DAM   &95.8 & 78.5 & 91.5 &98.8  &87.4 \\
CSRAN &96.0 &  79.8 & 92.5 & 99.0 &88.2\\
\hline
\multicolumn{6}{l}{Cross-lingual information retrieval model}\\
\hline 
ML-BERT &97.2 &  83.8 & 95.0 & 99.3 &90.8\\
POSIT-DRMM &95.4 &74.3   &90.7  & 99.3  &85.1\\
CLMN$^{-}$ &97.1 &83.5   &94.7  &99.6  &90.5\\
\bf CLMN &\textbf{97.8} &\textbf{86.8}   &\textbf{95.5}  &\textbf{99.8}  &\textbf{92.3}\\
\hline
\end{tabular}}

\caption{\label{tab:main} Evaluation results of baselines and our models. CLMN$^{-}$ refers to CLMN without using pre-trained BERT.}
\end{table}

\section{Experimental Results}

Table \ref{tab:main} summarizes the evaluation results of different cross-lingual retrieval models. 
Overall, our proposed CLMN model achieves the best performance on five evaluation metrics.
For simplicity, we mainly discuss the performance of $R_{10}@1$ in the following part.
In details, the advanced unsupervised CLIR models only achieve a fair performance without the collected CLIR-EC dataset. 
Through utilizing paired data, translation based CLIR models significantly outperform the SOTA unsupervised methods and achieves a decent performance on each metric.
We can observe from Table \ref{tab:main} that the attention-based translation CLIR model DAM achieves a noticeable improvement upon the SMN model, which demonstrates the effectiveness of attention module for modeling the unordered product attribute sets.
Moreover, directly modeling the cross-lingual information retrieval task can further introduce a significant performance improvement over the translation based models.
Both the recent proposed POSIT-DRMM and ML-BERT models yield a large increase over the strong translation based models. 
We also observe that our proposed CLMN$^-$ matching model are better than the other matching models and even achieves comparable results with the fine-tuned multi-lingual BERT.
With the enhancement of monolingual BERT, our proposed CLMN model outperforms the fine-tuned multi-lingual BERT model.

\subsection{Ablation Study}
Table \ref{tab:ablation} presents the experimental results for ablation study. 
Through analyzing, we observe that monolingual BERT is superior to train domain-specific word embeddings in our task.
All models trained on the monolingual BERT achieve impressing results in Table \ref{tab:ablation}, which are significantly better than domain-specific word embeddings pre-training.
We attribute such an observation to the relatively small size of the domain-specific monolingual data, which is far from ideal to train a decent model.
As for the influence of cross-lingual alignment, it can effectively capture the correlations between two different languages and thus can introduce performance improvement.
Besides, we also notice that the context-dependent mapping strategy is more effective in English bilingual space.
As reflected in Table \ref{tab:ablation}, CLMN-ch2en model yileds a observable improvement over CLMN-en2ch model on two different settings.
This observation is perhaps owing to that the relatively small size of the unordered Chinese product attribute can be easily mapped to another language space compared with a relatively long product description with lexical and syntactic structures.

\begin{table}[!t]
\centering
\large
\resizebox{\columnwidth}{!}{
\begin{tabular}{|l c c c c c|}
\hline
\textbf{Model}   & \bm{$R_2@1$}  & \bm{$R_{10}@1$}  &\bm{$R_{10}@2$} &\bm{$R_{10}@5$} &\bm{$MRR$}\\
\hline 
\multicolumn{6}{l}{Initialize with word embedding}\\
\hline 
CLMN$^{-}$-mono &95.8 &78.3   &91.6 &99.2  &87.3\\
CLMN$^{-}$-en2ch & 96.3 &81.5 & 93.1 & 99.5 & 89.2\\
CLMN$^{-}$-ch2en &96.7 &81.8   &93.4  &99.5  &89.5\\
CLMN$^{-}$ &97.1 &83.5   &94.7  & 99.6  &90.5\\
\hline 
\multicolumn{6}{l}{Initialize with BERT}\\
\hline 
CLMN-mono &96.5 &80.7  &93.8  &99.8  &89.1\\
CLMN-en2ch & 97.2 & 84.6 & 95.4 & 99.7 & 91.4\\
CLMN-ch2en &97.5 &85.5   &94.8  &99.8  &91.7\\
\textbf{CLMN} & \textbf{97.8} & \textbf{86.8} & \textbf{95.5} & \textbf{99.8} & \textbf{92.3}\\
\hline
\end{tabular}}
\caption{\label{tab:ablation} Ablation study of cross-lingual alignment, where \textit{mono} refers to without cross-lingual alignment learning, \textit{en2ch} refers to calculating the matching score in the Chinese bilingual space, and \textit{ch2en} is to compute the relevance score in the English Bilingual Space.}
\end{table}

\begin{figure}[t]
\centering
\includegraphics[width=1\columnwidth]{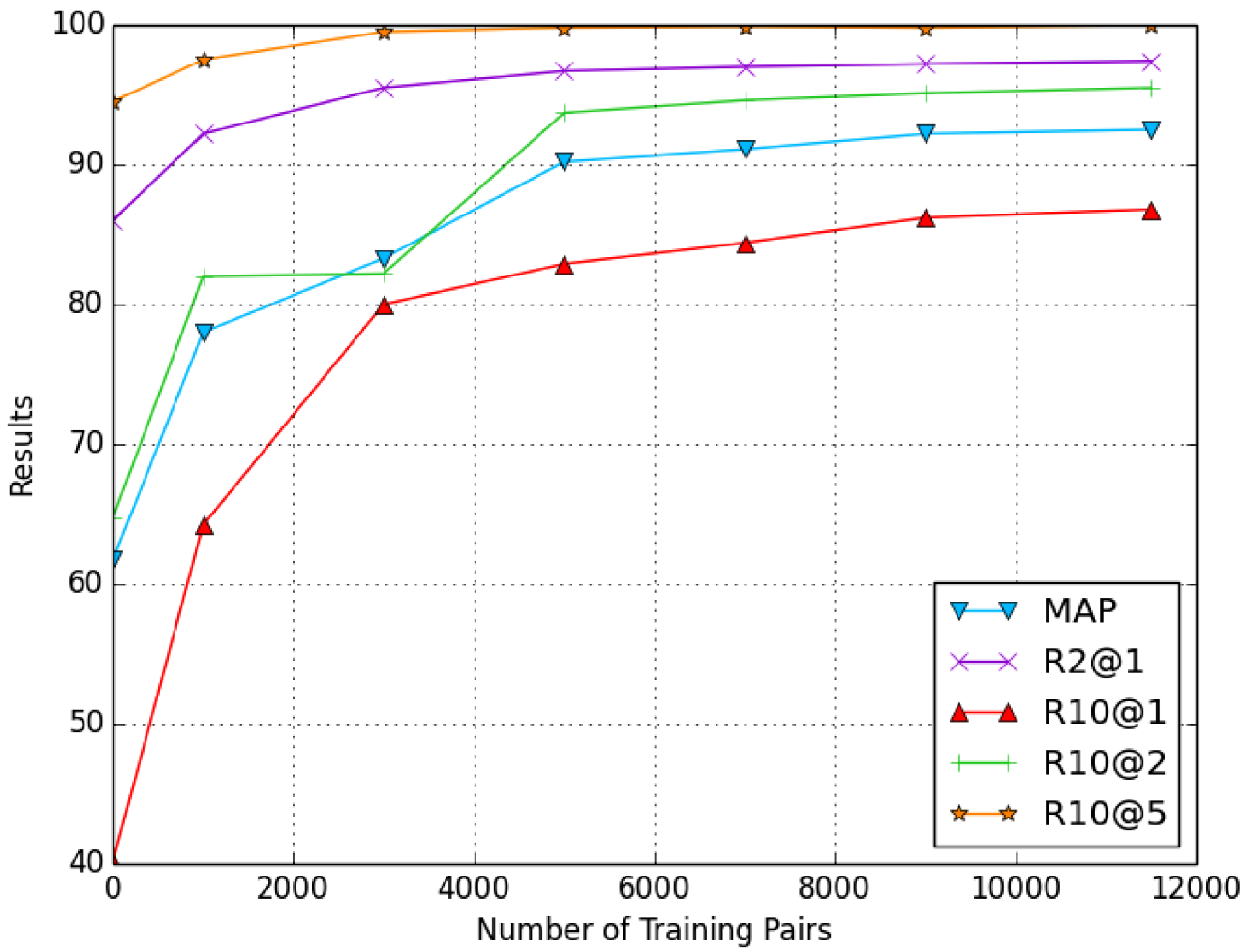}
\caption{\label{fig:size}
Trade-off between the size of CLIR-EC corpus and model performance.}
\end{figure}

\subsection{The Effect of Paired Data Size}

\begin{figure}[t]
\centering
\includegraphics[width=1\columnwidth]{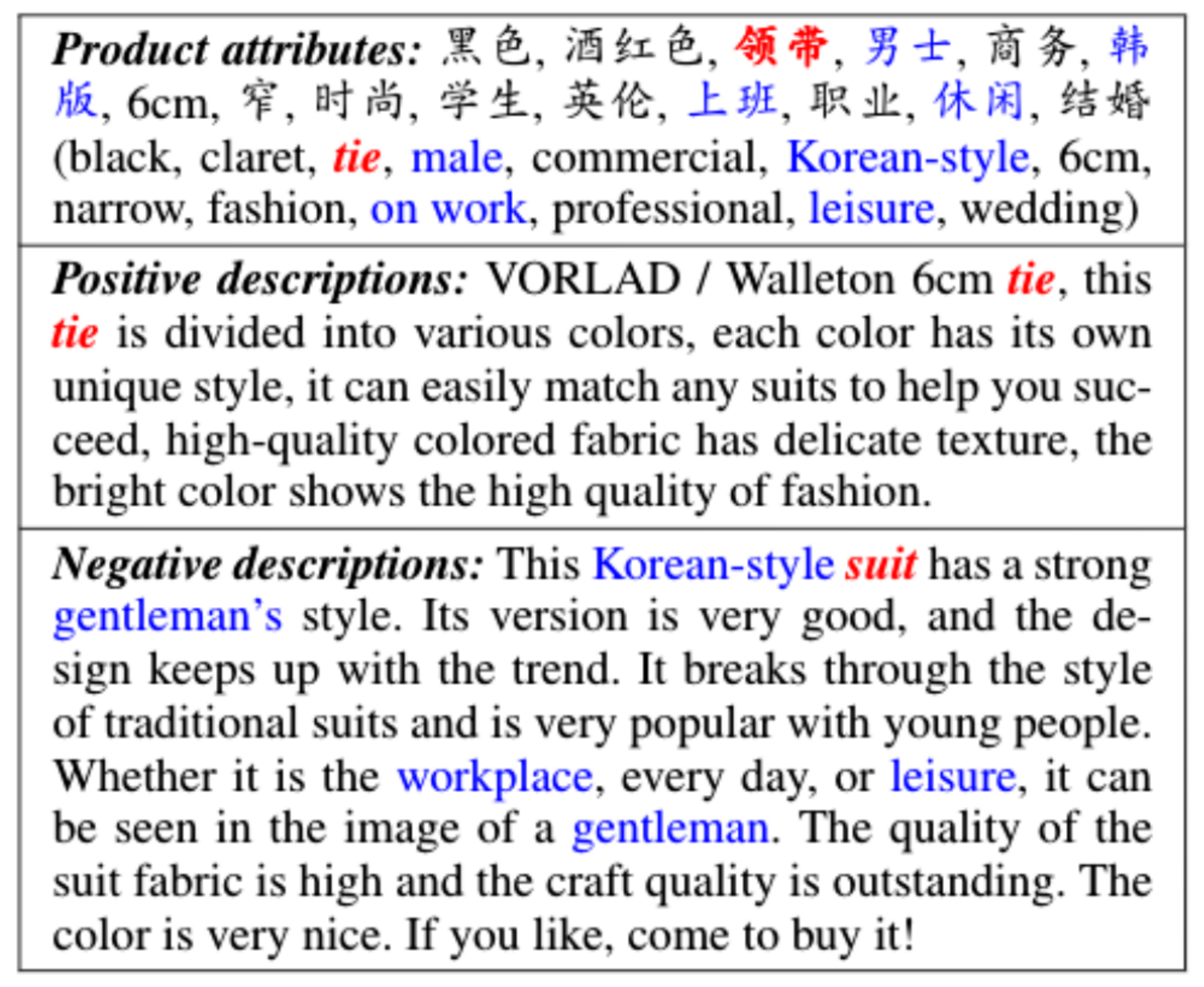}
\caption{\label{tab:bad}A typical bad case example in our task that the proposed CLMN model fails in $R_{10}@5$ .}
\label{fig:bad_case}
\end{figure}

We also launch experiments to study the influence of the pried data size for directly cross-lingual retrieving. 
As illustrated in Figure \ref{fig:size}, when the number of training pairs is less than 4000, the performance of cross-lingual information retrieval is positively correlated with the increasement of cross-lingual training pairs. 
When the number of cross-lingual training pairs is large than 4000, model performance only gains a slight improvement with more training pairs.

\subsection{Bad Case Analysis}
To learn the limitations of our proposed CLMN model, we thoroughly analyze typical bad cases from the test set that fails in $R_{10}@5$.
Figure \ref{tab:bad} presents the typical example that our proposed model fails in the cross-lingual retrieval task.
We can observe that the main error pattern is caused by the commonly occurred phrases in e-commerce platforms, such as ``gentleman", ``leisure", ``Korean-style", ``workplace", ``male".
The system pays more attention to these words and leaves the key information unobserved, e.g., the trained model retrieves a product description of "suit" for the product of "tie". 
Such an error pattern is related to the search behaviors of customers on an e-commerce platform.
Usually, users are not exactly sure about their target products.
They prefer to input a large concept such as ``man", ``Korean-style", rather than ``6cm black tie".
To improve the search conversion rate, sellers on e-commerce platforms add these general keywords to their product attributes.
As a result, the potential solutions of this issue rely on how to filtrate this general information and augment the matching signals of important product attributes.

\section{Conclusion}
In this paper, we propose to address the task of cross-lingual set-to-description retrieval in e-commerce, which involves automatically retrieving a persuasive product description in the target language given a product attribute set in the source language.
We collected a high-quality heuristic dataset with 13.5K pairs created by experts and two large monolingual auxiliary corpora for sharing.
We further propose a novel cross-lingual matching network (CLMN) aligned with a refined context-dependent cross-lingual mapping upon pre-trained BERT to address this low-resource cross-lingual retrieval task directly.
Experimental results indicate that our proposed CLMN is effective for solving this challenging task.
In our future work, we will evaluate our model on other open datasets for cross-lingual information retrieval.

\section{ Acknowledgments}
We thank the reviewers for their constructive comments.
This work was supported by the National Key Research and Development Program of China (No. 2017YFC0804001), the National Science Foundation of China (NSFC No. 61876196 and NSFC No. 61672058). Rui Yan was sponsored by Alibaba Innovative Research (AIR) Grant and Beijing Academy of Artificial Intelligence (BAAI).

\bibliographystyle{aaai.bst}
\fontsize{9.3pt}{10.2pt}\selectfont
\bibliography{7441-aaai.bib}

\end{document}